# Language Models as a Knowledge Source for Cognitive Agents

**Robert E. Wray**                                                                 ROBERT.WRAY@CIC.IQMRI.ORG
**James R. Kirk**                                                                   JAMES.KIRK@CIC.IQMRI.ORG
**John E. Laird**                                                                    JOHN.LAIRD@CIC.IQMRI.ORG
Center for Integrated Cognition@IQMRI, 24 Frank Lloyd Wright Dr., Ann Arbor, MI 48105 USA

## Abstract

Language models (LMs) are sentence-completion engines trained on massive corpora. LMs have emerged as a significant breakthrough in natural-language processing, providing capabilities that go far beyond sentence completion including question answering, summarization, and natural-language inference. While many of these capabilities have potential application to cognitive systems, exploiting language models as a source of task knowledge, especially for task learning, offers potential for significant, near-term benefits. We introduce language models and the various tasks to which they have been applied and then review methods of knowledge extraction from language models. The resulting analysis outlines both the challenges and opportunities for using language models as a new knowledge source for cognitive systems. It also identifies possible ways to improve knowledge extraction from language models using the capabilities provided by cognitive systems. Central to success will be the ability of a cognitive agent to itself learn an abstract model of the knowledge implicit in the LM as well as methods to extract high-quality knowledge effectively and efficiently. To illustrate, we introduce a hypothetical robot agent and describe how language models could extend its task knowledge and improve its performance and the kinds of knowledge and methods the agent can use to exploit the knowledge within a language model.

## 1. Introduction

Recently, there have been remarkable advances in natural-language AI capabilities. Specialized, deep pretrained language models, such as GPT-2 (Radford et al., 2019), GPT-3 (Brown et al., 2020), and Turing-NLG (Microsoft, 2020), are sentence completion engines. They are trained on massive corpora of language to predict the next word(s) given a prompt. After training, they can be fed a text prompt and they then generate a sequence of words that completes the text prompt. Language models have demonstrated an almost eerie ability to (sometimes) create coherent text that suggests a deep understanding of a subject, even though they lack internal model(s) of the world. Language models offer potential utility beyond language generation alone. They are setting new performance benchmarks across a range of familiar NLP tasks, such as question answering, common sense inference, and sentiment analysis (Bommasani et al., 2021).

These language models have not yet been integrated with more complete cognitive agents (to our knowledge). Today, language models are used as standalone systems or components of systems specialized for a single task, such as chatbots (Adiwardana et al., 2020). While language models can enhance human-agent natural language interaction, this paper emphasizes the exploitation of large-scale language models (such as GPT-3) as one source of knowledge for cognitive agents that learn new tasks and specialize their execution of tasks for individual contexts. Ready access to new knowledge is a long-standing barrier for practical cognitive systems. A recurring research thread over the history of cognitive systems has focused on various methods for extracting new knowledge from external sources (including humans) and then using that extracted knowledge to improve behavior and/or to perform new tasks (Forbus et al., 2007; Gluck & Laird, 2019; Hinrichs & Forbus, 2012; van Lent & Laird, 2001; Wray et al., 2004). Drawing on this history and foundation, we view language models as a potential source of knowledge to be exploited by cognitive systems, somewhere between the extremes of full natural language understanding (e.g., reading) and highly curated knowledge bases. Cognitive systems also may contribute capabilities and tools to improve the practical utility of knowledge extraction from language models.

While there is current research seeking to extract structured knowledge from language models (reviewed below), there are no established methods that would assure the provenance and relevance of the extracted knowledge given the uncurated nature of language-model source data and non-determinism in results from extraction. Ongoing research and investment in language models essentially guarantees continual improvements so that understanding how to exploit them as knowledge sources for cognitive agents also has potential benefits now and into the future.

To illustrate how language models might be used with a cognitive system, we introduce a hypothetical robot task, inspired by our previous work in interactive task learning (Kirk, 2019; Laird et al., 2017; Mininger, 2021). The robotic task is to support humans in either a household, office, or warehouse environment. These similar but distinct task contexts illustrate differing needs of a robotic agent and how different contexts can influence (for better and worse) what a language model can produce. While the examples are specific in order to ground discussion in specific needs and requirements, our goal is to identify general techniques and potential capabilities (and challenges) for extraction of knowledge from language models that are generally applicable to cognitive systems research.

We emphasize both what cognitive architectures do best (support end-to-end integration of interaction, reasoning, language processing, learning, etc. using structured, curated knowledge) and what language models do best (provide associational retrieval from massive stores of latent unstructured, possibly unreliable knowledge). Based on the analysis presented in this paper, we conclude that successful utilization of a language model will require a cognitive agent to develop its own internal model of how to use language models to 1) characterize the availability and quality of different domains and types of knowledge implicit in the language model, and 2) characterize the most effective techniques for extracting different domains and types of knowledge implicit in the language model.

## 2. Language Models and Their Uses

Current language models (LMs) generally have an encode, decode, or encode-decode architecture, as represented in Figure 1. This architecture can classify and/or produce sequences from sequences (Sutskever et al., 2014), which is particularly apt for natural language processing given the sequential nature of language. The encode component (a deep neural network) learns a *context vector,* a representation of individuals words (or *tokens*) in the context of sequences in which they appear in a training set; the LM *auto-encodes* this context via presentation of sequences during training. The encode process is bi-directional, meaning that tokens that come both before and after a particular token influence the context vector that is learned for a given word (both "robot" and "box" inform the encoding of the "move" feature in the context vector). Auto-encoder LMs are generally used for text classification, such as sentiment analysis, grammar classification, or word labeling (e.g., entity type, part of speech).

The decode component produces outputs that predict the next item in a sequence. Decoding is unidirectional; only words that come before (or after, but not both) a particular word influence production of that word.

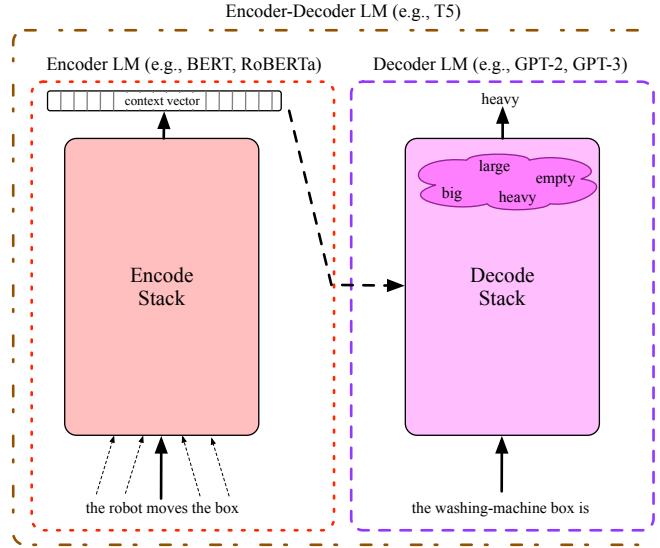

*Figure 1*. High-level architectures of language models.

Decoding is *autoregressive*, meaning that when a token/word is produced as output, that generated token can then influence the production of a subsequent one. Auto-regression thus allows a chain of tokens to be produced from a fixed input sequence. Decoders are the engines of sequence/sentence completion in LMs and are used for natural language generation (NLG).

As suggested by the figure, a particular LM can include an encoder stack, a decoder stack, or both. Encoder-decoder LMs, such as T5 (Raffel et al., 2020), are used for sequence-to-sequence transformations, such as language translation and summarization.

Language models were made more practical by the emergence of the transformer architecture (Vaswani et al., 2017). Transformer networks are easier to parallelize than recurrent neural networks resulting in faster training with GPUs. They also enable a deep network to direct attention to whatever part of a long sequence is relevant to the current state. This self-attention enables long-range dependencies across different parts of a sequence to inform both encoding and decoding.

Transformer-based LMs have made it practical to train large-scale models with billions of parameters. While it is possible to train a LM de novo, significant investment is required. Consequently, most researchers and developers today use models pre-trained on some text corpus. We use "language model" as a short-hand for transformer-based, pre-trained language models. Examples include BERT (Devlin et al., 2019) and derivative LMs such as RoBERTa (Y. Liu et al., 2019), the GPT family (Brown et al., 2020; Radford et al., 2019), Turing-NLG (Microsoft, 2020), Switch Transformer (Fedus et al., 2021), and numerous others (Bommasani et al., 2021).

Training a LM does not require labeled instances, which were a significant barrier to large-scale natural language (NL) systems a decade ago. Instead, language modeling is an unsupervised (or *self-supervised*) learning task in which the system learns an estimated probability distribution for completing the next token in a prompt sequence (e.g., a phrase or partial sentence) given a training set of such sequences (a corpus of natural language text). Auto-regressive decoders produce a chain of predictions that complete the input sequence. The input prompt to a pre-trained language model can include both a phrase to be completed and a context (in the form of words and sometimes special symbols), which biases a LM toward producing completions consistent with the context.

The number of parameters in the models have expanded rapidly in just a few years as it became apparent that performance improved with increasing scale, from 110M parameters in BERT (Devlin et al., 2019), to 175B in GPT-3 (Brown et al., 2020), to 1.6T parameters in the recent Switch Transformer (Fedus et al., 2021). Very large NL generation (NLG) models have surprisingly impressive capabilities given that what they are learning is probability distributions of tokens over previously seen sentences. Even a simple prompt can be sufficient for generating relevant text.

LMs have shown state-of-the-art performance on open-ended textual tasks such as predicting likely sentence completions from a sentence fragment (Melis et al., 2018) and many general language-tasks, such as text classification (e.g., sentiment analysis), translation, summarization, and question answering — without any specific pretraining or fine-tuning beyond the language prediction tasks (Y. Liu et al., 2019; Wolf et al., 2019).

Our primary interest arises because LMs encode not only language-usage knowledge but also world knowledge and commonsense knowledge (Bosselut et al., 2019; Forbes et al., 2019; Petroni et al., 2019). While this work is promising, the knowledge is clearly implicit in the parameters of the language model and it may be difficult to extract reliably (Cao et al., 2021; Mostafazadeh et al., 2020). Thus far there appears to be only nascent progress in systematic evaluation of the knowledge that language models contain and can express (Davison et al., 2019). Further, some aspects of task knowledge are known to be particularly difficult to transfer (or to extract) because such "tacit" knowledge is unique to an individual and often cannot be readily conveyed in language (Ambrosini & Bowman, 2001; Polanyi, 2009). Given these potential challenges, from the cognitive systems perspective, an open question is whether LMs can be exploited to extract useful world knowledge, commonsense knowledge, and language knowledge for use by a cognitive agent.

## 3. Challenges and Opportunities for using LMs in Cognitive Systems

Many cognitive systems need to learn new tasks but acquiring knowledge efficiently has been a persistent roadblock for practical success with cognitive systems. Interactive task learning (ITL; Laird et al., 2017) has shown that it is feasible for agents to learn new tasks via human interaction, but that interaction can be tedious for the human user as the agent requires that everything be explained. LMs are a source of knowledge that can potentially speed up the ITL process and reduce burden on the human instructor. While there are also opportunities to use LMs to improve the naturalness of human-agent interaction, in this paper we emphasize techniques to enable ITL systems to extract knowledge from LMs during and in service of task learning.

We define *knowledge extraction* as the process by which an agent gains knowledge of its task and/or environment from an external knowledge source. Thus, successful extraction results in the agent having (new) knowledge it can bring to bear on its tasks. From a cognitive-systems perspective, what is important is that the knowledge produced by extraction is "actionable." Thus, the goal of extraction is not simply to add knowledge, but to add knowledge that results in an agent improving its ability to function as an autonomous entity in a multitask environment.

While LMs promise to be a potential knowledge source, they will be challenging to use within a cognitive agent. We characterize features of LMs, both positive and negative, related to knowledge extraction in a cognitive agent. We contrast them with traditional knowledge sources, such as an internal semantic memory or external, curated knowledge bases, such as Cyc.

- **Breadth and Depth of Knowledge**: A major strength of LMs, such as GPT-3, compared to curated knowledge bases (KBs) is the extensive breadth of knowledge encoded within them.
- **Provenance and Accuracy of Knowledge**: The quality of behavior for a cognitive agent is invariably tied to the quality of knowledge it reasons with. With traditional KBs, the knowledge is either curated or at least derived from the agent's own experience with the world. In contrast, LMs are (largely) derived from uncurated text corpora. The provenance of the knowledge is unknown and very likely includes errors and conflicts.
- **Relevance of Knowledge**: Even when the LM contains knowledge relevant to the agent's needs, extracting it can be highly sensitive to the specific context and prompts used to elicit responses from the LM. This sensitivity to the relevant context makes it difficult to ensure that whatever information is retrieved is actually relevant to the target context of the agent.
- **Temporal Currency of Knowledge**: LMs are a prisoner of the data that was current at the time of training, and it is not feasible to continually retrain them with the most up-to-date information. Thus, they are ignorant of current events and any knowledge derived from recent events, such as the current president of the United States, or even the existence of the COVID pandemic. Therefore, a significant challenge for cognitive agents is to ensure that whatever knowledge they extract from an LM is accurate in the present time.
- **Situatedness of Knowledge**: Curated knowledge bases (such as Cyc) and LMs encode knowledge about the world, and even specifics of historical events and objects (such as the location of the Eiffel Tower). However, none of these sources encode knowledge about an agent's current situation, embodiment, goals, and plans, which may be encoded in an agent's long-term semantic or episodic memory. Thus, the knowledge that could be extracted is limited to general knowledge, not situational knowledge.
- **Model of Knowledge**: In using a traditional knowledge base, an agent knows in general its scope, reliability, level of detail, etc. A weakness of LMs is that it is not clear what knowledge is available, what form it is in, how accurate or dependable it is, and so on.
- **Accessibility of Knowledge**: In a typical AI KB, the APIs for query/response and knowledge representation are well defined, making it straightforward for an agent to attempt to retrieve information and to parse any responses. For a LM, the specific form of a request and response are (generally) less structured, e.g., sentences (sequences of tokens). Thus, a significant challenge for an agent will be to interpret results, where, in the extreme, an agent must parse natural language to extract what information is provided by the LM.
- **Structural Integration**: Traditional knowledge bases have low to moderate computational costs and latency and have reliable access. In contrast, many LMs, especially the largest, are web resources with restricted access, high relative latency whose access depends on internet connectivity. A cognitive agent will thus need to be strategic in using an LM in applications that involve real-time environmental and human interaction.

To achieve actionable knowledge extraction, a number of immediate issues must be addressed. The issues of *Accuracy* and *Relevance* are probably the most significant challenges. The agent needs methods that allow it to elicit the LM to produce highly relevant responses (when it can potentially produce almost anything). In response, we identify targeted prompting strategies that we hypothesize will mitigate sensitivity (§5.2.2). Because accuracy and relevance of returned results are potentially inaccurate, the agent will also need methods to test and to evaluate results (§5.2.4). Because all of these strategies will depend on the characteristics of specific LMs and will likely need to be refined and tuned with experience, the agent will require a *Model of Knowledge* (§5.2.1) to support these functions.

Additionally, for most of the existing extraction methods we review (§4), knowledge extracted from a LM is in the form of natural language ("the large box is empty"), highlighting the lack of immediate *Accessibility* of results. Thus, an agent will also require additional capabilities to turn a response phrase into knowledge it can use (§5.2.3)

## 4. Current Approaches to Extracting Knowledge from Language Models

In this section, we identify and review four existing approaches to knowledge extraction from LMs. Via initial discussion of strengths and weaknesses, we begin to assess which methods are most likely to be transferable to use in cognitive agents. Section 5 draws on this review and outlines the extraction process we envision to be most useful for cognitive systems.

Current approaches to knowledge extraction generally fall into four categories, some of which focus on identifying an apt context for extraction and some that seek to bias or tune the LM itself to the extraction task. The four classes of methods are:

1. Fine-tuning LMs for specific extraction tasks
2. Simple prompting using masked language models
3. Contextual prompting
4. Analogical or case-based prompting

### 4.1 Fine-tuning for specific extraction tasks

Fine-tuning customizes a pretrained LM through further training. Fine-tuning can refer both to tuning the content of the LM itself (e.g., introducing specialized vocabulary and language context, such as medical corpora) or tuning the model to emphasize a particular language task. Researchers have used task-focused fine-tuning to enable direct knowledge extraction from LMs. For example, COMET (Bosselut et al., 2019; Hwang et al., 2021) introduces labeled commonsense statements in the form of subject (a short phrase: "taking a nap"), relation ("causes"), and object ("increases energy") to fine-tune a pretrained LM. After fine-tuning, the LM can "fill in" unspecified objects when presented with a subject and relation, so that the model generates a predicted object for a given subject and relation (e.g., prompted with **sub(taking a shower), rel(causes)**, COMET will respond with an object such as "*being clean*").

Knowledge extraction via this task-based, fine-tuning process results in novel assertions that that humans find plausible and can thus be added to a compatible knowledge store, thus extending the knowledge available from that resource (Bosselut et al., 2019; Hwang et al., 2021). However, there are two significant limitations of the fine-tuning approach. First, fine-tuning is targeted for a single, well-defined extraction task. The agent may have many distinct extraction tasks and some of these tasks may not be so clearly definable in advance. As a consequence, an agent might require many distinct fine tunings, one for each task, which may be both computationally impractical and make the agent unresponsive to its environment (*Structural Integration*). Second, even if fine-tuning could be generalized to cover all agent extraction tasks, for very large-scale LMs, such as GPT-3, fine-tuning can require large computational resources for large datasets. As we discuss further below, fine-tuning may be apt for extraction tasks that recur across tasks.

## 4.2 Simple, mask-based prompting

Auto-encoder models are often used as masked language models (MLMs) that produce a specific, predicted word based on an input prompt. MLMs provide "fill in the blank" functionality in response to a prompt. For example, given the prompt "**A robot can <mask> a package**" RoBERTa (Y. Liu et al., 2019) responds with predicted completions, such as *open*, *deliver*, *inspect*, *carry*, *assemble*.[1] One of the first examples of knowledge extraction from LMs used MLMs to extract factual knowledge from *cloze* (fill-in-the-blank) sentences (Petroni et al., 2019).

Prompt construction for MLMs is straightforward from a syntactic perspective (a sentence with a mask), but creating an appropriate prompt requires relevant semantic knowledge. For example, to construct the prompt in the example above, an agent needs to know that robots and packages were relevant to the knowledge that it desired. The masked cloze prompts in the study mentioned above were manually created (Petroni et al., 2019). One of the emerging critiques of MLM-based prompting is that most of the "knowledge extraction" in this work is being done by the prompt rather than the language model (Cao et al., 2021). For a cognitive system using MLMs, the core issue is whether it can construct a prompt sufficient to provide needed knowledge.

## 4.3 Context-based prompting

Context-based prompting is an extension of MLMs in which a system learns to create a prompt rather than relying on manually created prompts. For example, AutoPrompt (Shin et al., 2020) uses gradient descent on labeled training data to find how to best prompt a LM model for a given NLU task (e.g., sentiment analysis, fact retrieval), learning via a search over the LM itself. AutoPrompt results are comparable to manual prompting. However, the best-performing prompts learned by AutoPrompt are not intuitive and differ substantively from prompts created by humans. For example, consider a prompt designed to elicit the position an athlete plays on a team. The manually authored relation is "<subject> plays in <mask> position" in (Petroni et al., 2019). This prompt is human interpretable, and one can anticipate possible substitutions for the mask. The prompt learned by AutoPrompt is: "<subject> ediatric striker ice baseman defensive <mask>" (Shin et al., 2020).

Another approach uses existing, structured knowledge to inform automatic construction of prompts (Bian et al., 2021). Externally derived, context-based prompting provides results comparable to AutoPrompt and manual prompting, but the prompt construction process requires less search over LM parameters and is more readily interpretable by humans. Because a cognitive agent will have structured knowledge on which to draw for prompt construction (e.g., its semantic and episodic memories), this approach may be able to take advantage of an agent's prior experience in combination with the task context to enable agent creation of effective prompts to an MLM.

Another example of context-based prompting is to use generative models to complete a statement (Table 1). In this case, the LM completes the sentence that is started by the prompt.

---

[1] In this and further LM examples, the prompt sent to the LM is highlighted in **bold**, the text generated by the LM is *italicized*, and the LM used to generate the example is specified in [square brackets] if not already identified in the text.

Although it may be possible to construct a prompt that leads GPT-2 to generate household furniture items, the straightforward prompt here is not sufficient. In contrast, GPT-3 generates a list that appears responsive to the original prompt.[2]

*Table 1*. Comparison of the responses of GPT-2 and GPT-3 to same/similar prompts.

| LM Prompt | LM Response |
|---|---|
| **Many household furniture items are stored in the warehouse including** | *household towels* [GPT-2] |
|  | *beds, tables, chairs, wardrobes, and more* [GPT-3] |
| **Many furniture items are stored in the warehouse including** | *tables, chairs, and other items* [GPT-3] |

Recent research has shown that GPT-3 prompts that are semantically similar to the target perform better than other prompting strategies (J. Liu et al., 2021). In the table, for example, "household furniture" has a close semantic relationship to "beds," "tables," and "chairs." In the third example in the table, removing "household" from the prompt results in furniture items that are common in non-household contexts (like an office). This responsiveness to the context in GPT-3 suggests that cognitive systems may have a simpler route to knowledge extraction via the use of larger-scale models. The agent can likely use what it already knows about a task or situation as context for a prompt, rather than using exhaustive exploration of the options such as is needed for the MLM approaches. However, the agent will need a model of the sensitivity of GPT-3 to specific forms of prompts and knowledge about how to manipulate GPT-3 parameters to utilize this method.

### 4.4 Analogical/case-based prompting

Large-scale, generative LMs make a new category of prompting possible, in which examples similar in form to the desired response are embedded within the prompt, creating an analogical context for the core question. A feature of GPT-3 is its ability to apply analogical cases to generation (Brown et al., 2020), and, as suggested by variety of examples in Table 2, GPT-3 can take advantage of syntactic indicators (such as "Q" and "A") to further refine its response. This approach has limits and the generalization from cases is narrow. Research suggests that the results from analogical prompting are attributable to setting expectations for entity-types for a response (Cao et al., 2021) and generalization of cases may perform differently in similar-seeming situations.

Even with limitations, this approach appears relevant and potentially beneficial to cognitive systems. Many of the other approaches introduce a constraint in terms of bootstrapping: the agent needs to have specific prior task/knowledge context in order to construct a good prompt. The potential advantage of analogical prompting is that an agent can draw on past experiences and potentially related prior task knowledge to create a prompt designed to draw out responses focused on a new task about which it knows little.

---

[2] GPT-3 examples, other than where noted, were generated with its temperature parameter=0, meaning that the response is largely deterministic/repeatable.

*Table 2*. Examples of analogical prompting using GPT-3.

| Prompt | Response |
|---|---|
| **The household robot charges in the garage.** **The office robot charges in the maintenance closet.** **The warehouse robot charges in the** | *storage room* |
| **When a room is occupied, the robot should not adjust the temperature or lights.** **When an office building is unoccupied, the robot should turn the heat to 60 degrees.** **When the warehouse is unoccupied, the robot should** | *turn off the lights* |
| **Q: Where should a robot find a package in an office?** **A: the mail room** **Q: Where should a robot find a package in a warehouse?** | *A: the shipping department* |

## 5. A Cognitive-systems Approach to Knowledge Extraction from LMs

We now outline a knowledge-extraction process from LMs for cognitive agents. Cognitive agents routinely interact with external knowledge sources to obtain new knowledge. The approach builds on familiar steps (and associated capabilities) in that process. However, as described above, interacting with LMs introduces a number of challenges to extract knowledge successfully and thus provide agents with actionable knowledge that improves their ability to perform new tasks. We identify and describe four research questions and plans to address them in the near-term:

1. Characterizing LM usage
2. Prompting strategies for acquiring task knowledge
3. Interpreting responses from an LM
4. Verifying the knowledge obtained from a LM

As an illustrative example of a cognitive agent that could benefit from a LM, we use Rosie (Kirk, 2019; Mininger, 2021), an existing interactive task learning agent implemented in Soar and embodied in multiple robots and simulation environments. While we have not yet integrated a LM with Rosie, we use actual examples from GPT-3 in the previously introduced household, office, and warehouse robotic domains to illustrate possible use cases and relate them to Rosie's capabilities.

### 5.1 Steps in Acquiring Agent Knowledge from a LM

*Figure 2* presents a general processing strategy for a cognitive agent to extract knowledge from a LM. Rather than extracting knowledge as a general goal, the cognitive agent's goal is to identify knowledge for a specific purpose: enabling the agent to perform a task. The extraction strategy is designed to take advantage of prior work in cognitive systems in accessing and using external knowledge sources (some of these steps are used for internal knowledge access as well). Steps that overlap with recurring patterns are illustrated in green; process steps specific to LMs are in blue.

In addition to the extraction process itself, we plan a complementary, overarching process (top arrow) that observes and evaluates interactions with specific LMs over the extraction steps (diamond lines) to construct a usage model for LMs. This usage model will explicitly encode meta-data that summarizes how to use the LM for specific questions and situations. The resulting model (represented by the database icon) is used by the steps in the extraction process to influence and inform how those steps are executed (dashed, open arrows).

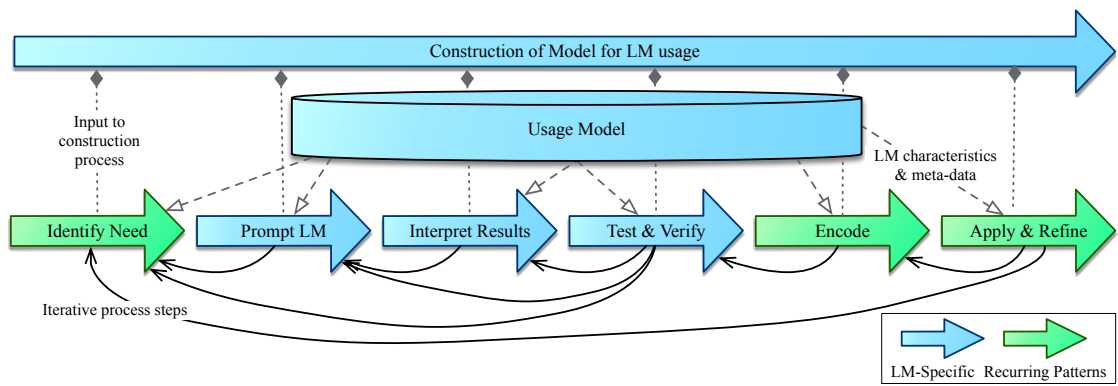

*Figure 2.* A step wise process for extracting knowledge for task learning in a cognitive agent.

The extraction process consists of the following six steps:

1. The agent **identifies a knowledge need**, such a gap in its knowledge and decides that a LM is an appropriate resource for potentially fulfilling its need. The Usage Model informs both the decision to use a LM and the choice of a specific LM.
2. The agent **prompts a chosen LM**, choosing a specific prompting strategy, creating the specific text for the prompt, and parameterizing the LM for the type of desired response. Ideally, the constructed prompt is also specific to the context, such as a particular type of knowledge gap. The Usage Model encapsulates how the agent should determine appropriate parameters for the LM.
3. The agent then **interprets the results** from the LM. Interpretation likely requires the agent to use internal natural-language understanding capabilities to convert the text to the agent's internal knowledge representation (§3).
4. Because the results from the LM are not necessarily accurate and reliable (§3), the **agent evaluates, tests, and attempts to verify extracted results** from the LM. Although verification of acquired knowledge has been explored in cognitive systems, verification of knowledge derived from LMs will be different in character (see below), and should be considered a new capability, hence its shading.
5. Following verification, the agent **encodes the knowledge it has obtained** into its own memory(-ies) as appropriate for current and future use in task performance.
6. In the final step, the agent **uses the knowledge** it has acquired and continues to monitor its correctness and utility, **and refines it** based on its experience in using it.

As suggested by the "iterative process step" arrows, this process will typically not proceed in a stepwise fashion but will be iterative with backtracking and restarts needed as the agent attempts to extract, test, and use accurate and relevant knowledge obtained from a LM.

### 5.2 Research Agenda

There are numerous research and engineering challenges that need to be overcome to extract knowledge from LMs, including revision of this initial conception of extraction process. However,

*Figure 2* provides an initial guide for pursuing extraction from LMs. In the context of this plan and the challenges introduced earlier (§3), we outline four specific research questions that we identify as critical to enable successful, actionable extraction.

### 5.2.1 Constructing a Model of LM Usage

What can kinds of knowledge can be readily extracted from language models (and a particular language model)? What parameters impact LM performance and how sensitive is the LM to those parameters? An agent will need to understand what LM characteristics are most salient to the challenge of extraction of agent knowledge from LMs and how those characteristics interact with the agent's prompting strategy. The Usage Model pictured in *Figure 2* is where that information will reside. The Usage Model will capture meta-data about the LM. As a simple example, meta-data should include the date when the LM was developed and what corpus was used to train the LM. This information can be used by the agent to help it decide if the LM contains desired information (e.g., if the agent needs to know the US president in 2021 and it has access to no LM trained as late as 2021, it needs a different source to answer to that question).

Long-term, as these questions are resolved and a concrete "track record" of LM performance is established, agents need to build up a characterization of a LM based on its own experience, using native learning and reasoning mechanisms to capture and to assess usage patterns. Today, understanding specific characteristics important to capture is not yet known to even researchers. Thus, we plan to focus on characterizing and understanding key performance dimensions via systematic exploration of various LMs applied to a task comparable to the robotic task. Systematic exploration will provide a clearer understanding of the properties of LMs in the context of their use and inform the enumeration of more fine-grained requirements for the Usage Model.

### 5.2.2 Prompting Strategies for Acquiring New Task Knowledge

The agent must choose a prompting strategy and the formulate a specific prompt appropriate for its task and environment, the knowledge need, and the requirements of the LM itself. We are exploring a two-pronged prompting approach, drawing on existing extraction approaches (§4).

**Analogical based prompting to explore a new task/problem space**: When an agent has little or no knowledge of a new task, it can be difficult for it to form targeted prompts for specific kinds of knowledge needs (below). However, using the analogical prompting approach provides a potential to extract initial information about a new task by attempting to compare it to a known task. The results of this kind of prompting will likely be more useful for supplying initial "hints" about the new task (e.g., examples of entities, concepts, and relations), rather than a complete specification. But even if this method provides only some initial lexicon for the task, it has the potential to enable more targeted prompting strategies.

**Template-based prompting for targeted concepts:** This approach draws from the context-based prompting approaches outlined above but also takes advantage of research in cognitive systems focused on learning specific kinds of knowledge. For example, the Problem Space Computational Model (PSCM; Newell et al., 1991) informs templates for certain classes of prompts, such as identifying the goals, objects, and relations associated with the task (state features), actions (operators), as well as guidance for choosing and comparing actions in certain states. Table 3 lists examples that we are currently exploring. An ellipsis is used for templates that would be used to prompt a generative LM, such as GPT-3, and <mask> is used to denote a prompt that sought to return a single word from a masked language model. *?item* indicates a variable (and variable type) in the template that is instantiated by the agent during prompt formation.

*Table 3. Examples of potential prompting templates inspired by the PSCM.*

| PSCM Functions | Illustrative Prompt (no context) |
| --- | --- |
| Problem space description | How do you… Explain … |
| Goal definition | What is the goal of ?task … |
| State definition (lexicon) | ?object is also known as a <mask> |
| State definition (taxonomic relations) | ?object is a type of <mask><br>?object has a/is part of a <mask> |
| Operator definition (lexicon) | ?actor can <mask> an ?object |
| Operator definition (preconditions) | ?object is used for <mask> |
| Operator definition (actions) | ?action causes… |

*Table 4* provides examples of the responses of various LMs to (manually) instantiated templates. Instantiating these templates requires substituting variables, but also can include adding context to give more information about the task. Context can be included by prepending relevant words to the prompt for masked language models or by including analogical cases for generative models (§4.4). As can be seen from the examples, the context can significantly change the response of the LM even when the template is the same.

This template-based approach may also be more amenable to fine-tuning. Such fine-tuning would be comparable to COMET's approach (as above). COMET attempted to extract many and varied relations covering a general knowledge base, such as ConceptNet (Speer et al., 2017). With this broad scope, its overall performance was good (agreement with human raters > 90%) but also sometimes generated nonsensical responses (Bosselut et al., 2019). However, for a template-based approach, fine-tuning would focus on the comparatively few and narrow relations defined by templates and thus might offer possibly greater reliability.

The GLUCOSE knowledge base (Mostafazadeh et al., 2020) provides evidence that fine-tuning for a well-defined, modest number of relations/rules is both technically feasible and produces more reliable results overall than over knowledge-extraction approaches. GLUCOSE is based on twelve relations relevant to commonsense inference in narrative understanding, such as "change in location" or "change in property" due to an action. Fine-tuning GPT-2 resulted in significantly better performance at extracting these relations from example stories than GPT-2 alone was able to do. The potential lesson for the use of template-based extraction is that fine-tuning based on the specific templates may significantly improve outcomes. Further, as long as the templates are not specific to a domain or task, fine-tuning may then be a more apt strategic tool for an agent to use. In this case, the agent would not need to fine-tune for each new task, which would likely be too computationally demanding to be practical.

While we have no expectation in the short-term that agents will extract a full problem-space specification from a LM (even piecemeal), this PSCM framing for the search for new knowledge seems likely to scaffold the acquisition of new task knowledge, as it has in earlier research focused on different knowledge sources (Ritter et al., 2006; van Lent & Laird, 2001; Yost, 1993).

*Table 4*. Examples of responses from instantiated, contextualized templates.

| Instantiated Template | LM Response |
|---|---|
| **Explain how to patrol a warehouse.** | *Walk around the warehouse to inspect it and make sure that it is free of people. Search the warehouse for dangerous items like weapons or hazardous materials.* [GPT-3] |
| **The goal of patrolling a warehouse is** | *to identify any hazards that may be present.* [GPT-3] |
| **A shelf is also known as a <mask>.** | *cabinet, rack, shelf, table, stand, drawer, platform, closet, trunk, box* [RoBERTa] |
| **A wall is part of a <mask>.** | *house, building, structure, home, wall, room, castle, prison, fortress, fence* [RoBERTa] |
| **A robot can <mask> a package.** | *open, deliver, inspect, carry, assemble, sort, transport, handle, check, sniff* [RoBERTa] |
| **A bay is used for <mask>.** | *navigation, fishing, shipping, boats, ships, sailing, transportation, storage, commerce, docking* [RoBERTa] |
| **warehouse robot. A bay is used for <mask>.** | *storage, loading, maintenance, equipment, docking, cargo, parking, recycling, deliveries, cleaning* [RoBERTa] |
| **Pushing a box causes** | *the box to move and the object inside the box to move with the box.* [GPT-3] |

*5.2.3 Interpreting the Responses of an LM*

A difficult and immediate challenge for agents using generative LMs is that they produce responses that are essentially natural language rather than easily parsed assertions from a knowledge base. For example, Rosie is designed to interact with humans and has language understanding capabilities, but they are limited to handling a restricted subset of NL. Near-term, we plan to use template-based prompting to limit the complexity of what is returned from the LM. This approach offers a lower utility from the agent-LM but is more tractable to support the initial research.

Another near-term approach is to use a LM to shape a prompt or to post-process the results with the goal to produce output consistent with the processing capabilities of the agent. Table 5 uses a prior agent dialogue with a user (from Rosie) that provides context in the prompt that influences how GPT-3 generates its results. The result with the prior dialogue is much more similar to dialogues Rosie has previously seen, increasing the likelihood that the agent can interpret them.

R. WRAY, J. KIRK, AND J. LAIRD### 5.2.4 Assessing and verifying results from the LM

Having obtained knowledge from the LM, the agent will need to evaluate if that knowledge is correct. Verification presents a significant challenge to agents using LMs. The results returned from cueing the language model may be unreliable (both due to *Accuracy* and *Relevance* challenges). *Table 6* provides an illustrative example. The first or last responses are probably the most useful one for a warehouse agent; the other two are likely incorrect (and at the very least, not relevant).

*Table 5*. Example of using GPT-3 to produce text more closely matched to agent NLU capabilities.

|  | Without prior dialogue | Including prior dialogue |
|---|---|---|
| Prompt | **Move the package into the cabinet. What is the next goal or subtask of move?** | **Move the box onto the table.**<br>**What is the next goal or subtask of move?**<br>**Pick up the box.**<br>**Put the box onto the table.**<br>**You are done.**<br>**Move the package into the cabinet.**<br>**What is the next goal or subtask of move?** |
| GPT-3 response | *The next goal or subtask is to move the package into the cabinet. Apply these steps to a goal or subtask until the lowest level of goal or subtask is reached.* | *Pick up the package.*<br>*Put the package into the cabinet.*<br>*You are done.* |

Research is needed to identify various ways that the agent can test and verify knowledge and how "potential knowledge" (knowledge that has not yet been fully tested and verified) can and should be used by the agent. That is, how can an agent evaluate the correctness of generated knowledge and how might it test and evaluate the knowledge before fully incorporating into its other memories and/or acting upon it? To enable near-term progress, we will employ feedback from a human (e.g., a dialogue with Rosie) to verify knowledge. Longer term, we will explore how an agent can perform this verification autonomously. Key questions include how an agent can measure confidence in its assessment of individual assertions and exploring how the agent can use its estimates of confidence to best balance its need for actionable knowledge and correctness in its knowledge.

*Table 6*. Example of the possible variation to the identical prompt using GPT-3.

| Prompt | Responses |
|---|---|
| **Normal operating hours for the warehouse is 6am-9pm on weekdays (closed on weekends). Staff should not generally be in the building once the 9pm shift ends. Staff will begin arriving just before 6. Because we are closed on the weekends, staff** | *should not be in the building on weekends* |
| | *can't arrive at the start of their shift* |
| | *typically do not arrive between 6-9 on the subsequent Monday* |
| | *will not be in the building on Saturday or Sunday* |

## 6. Conclusions

Via the analysis in this paper, we have outlined a research strategy that builds on current LM research as well as prior research in cognitive systems to attempt to specify effective patterns of prompting LMs, interpreting results, and verifying the resulting knowledge obtained from the LM. The analysis resulted in a roadmap for extracting knowledge from LMs that integrates concepts and methods from the way knowledge extraction from other sources has been implemented in cognitive systems and standalone methods for extracting knowledge from LMs. The analysis also identified four specific research challenges that will need to be addressed to realize knowledge extraction that can impact how agents learn and execute their tasks.

While both preliminary and far-reaching, this research is aimed toward one of the core needs of cognitive systems, the ability to learn new tasks, which generally requires rich sources of knowledge for efficient learning. As outlined in the analysis, it is not likely that LMs alone will provide a complete and robust source of knowledge for new learning new tasks. LMs will lack knowledge (e.g., limitations of the underlying corpus, tacit knowledge), retrieval and extraction will be imprecise, etc. However, in the context of long-lived cognitive systems that can learn new tasks, LMs offer the potential to complement and supplement other sources of knowledge. A LM is potentially more comprehensive than formal, curated knowledge bases, easier to understand and to encode in agent memory than text documents, and more readily and persistently available than a human. The analysis presented in the paper identifies specific research problems targeted to determining if language models can indeed be an effective bridge to new knowledge in cognitive systems.

Longer-term, as basic extraction processes mature, we see additional research questions related to LMs that are either specific to cognitive systems or likely have distinct requirements from current methods that use language models. These include:

1. **Adapting extraction for improved alignment**: Research in the LM community is already focusing on evaluating how knowledge in LMs conforms (and does not) with human understanding and how LMs can perpetuate social biases inherent in source corpora. Extraction processes should draw on this research to support agents in developing task knowledge that conforms to the expectations of users (alignment) and mitigates bias.
2. **Human-agent language interaction**: Natural language interactions, including comprehension and generation, are important for cognitive systems that work in human environments. Human speech is often not concise or even entirely grammatical, which makes it difficult for a cognitive system to parse or understand. LMs might help a cognitive system to classify new words, check grammar, evaluate semantic relatedness, etc., especially as *Structural Integration* issues are mitigated.
3. **Alternative patterns of LM integration**: We mapped the use of LMs to a systems-integration pattern that is used for accessing external knowledge stores. However, this pattern is one of several that may be worth considering, including integration of LMs as a component more like a memory in a cognitive architecture than an external store. Memory-like integration could simplify some aspects of LM (e.g., the usage model) although it could make others more difficult. However, as the community uncovers fundamental properties of LMs for use in cognitive systems, tighter integration patterns may be worth pursuing.

**Acknowledgements**

This work was supported by the Office of Naval Research, contract N00014-21-1-2369. The views and conclusions contained in this document are those of the authors and should not be interpreted as representing the official policies, either expressed or implied, of the Department of Defense or Office of Naval Research. The U.S. Government is authorized to reproduce and distribute reprints for Government purposes notwithstanding any copyright notation hereon. The authors thank Charles Newton of Soar Technology who provided suggestions and guidance on language models and the anonymous reviewers, who provided incisive feedback and suggestions, including the recommendation of additional research relevant to this work.

LANGUAGE MODELS AS A KNOWLEDGE SOURCE FOR COGNITIVE AGENTS

R. Wray, J. Kirk, and J. Laird